\documentclass[letterpaper, 10 pt, conference]{ieeeconf}
\IEEEoverridecommandlockouts
\usepackage{xcolor}
\usepackage{graphics}
\usepackage{epsfig}
\usepackage{booktabs} 
\usepackage{amsmath}
\usepackage{amssymb}
\usepackage{float}
\usepackage{stfloats}
\usepackage[normalem]{ulem}
\usepackage{setspace}
\usepackage{cite}
\usepackage{algorithm}
\usepackage{algpseudocode}

\title{\LARGE \bf Interpretable Collision Avoidance and Feasibility Guarantees through an Uncertainty Aware Pipeline}

\author{Doris Xu, Jin Hyun Park,  Guadalupe Alvarado Fernandez, Adhyan Prasad, Mark Campbell
\thanks{This work was supported by NSF Grants CNS-2211599, IIS-2305532}
\thanks{The authors are with the Sibley School of Mechanical and Aerospace Engineering at Cornell University, Ithaca, NY, 14850.}}

\begin{document}
\raggedbottom
\maketitle

\thispagestyle{empty}
\pagestyle{empty}

\begin{abstract}
Autonomous vehicles (AVs) operating in dynamic, challenging conditions such as intersections and roundabouts must guarantee collision avoidance under uncertainty while maintaining real-time feasibility. Most existing planners struggle to simultaneously guarantee both requirements.
Chance-constrained formulations embed safety constraints directly in the optimization but become infeasible as uncertainty compounds; deterministic planners provide feasible solutions but fail to address the uncertainty that causes collisions; end-to-end learned policies provide neither guarantee in an interpretable form.

We present a modular perception-prediction-planning pipeline that closes this gap by propagating uncertainty as structured, interpretable quantities across module boundaries. Rather than embedding probabilistic risk directly in the optimization, our pipeline resolves collision probabilities upstream and constructs convex safe sets within which the planner provides a guaranteed feasible solution at every cycle — decoupling the safety guarantee from solver feasibility. 

Across Monte Carlo trials in a four-way unsignalized intersection, our pipeline achieves 0\% collisions and 0\% solver infeasibility at 12.7ms per cycle, compared to 36\% and 10\% collision rates for deterministic and chance-constrained baselines respectively. The ablation study confirms the necessity of each uncertainty source and the hardware deployment on Clearpath Jackal UGVs validates the pipeline's real-time viability on physical platforms.

\end{abstract}
\section{Introduction}
\label{sec:Introduction}

Intersections account for approximately one quarter of all U.S.\ traffic fatalities annually~\cite{fhwa_intersection}, presenting some of the hardest challenges for autonomous driving. When considering autonomous vehicle pipelines, there are a number of key challenges. First, sensor measurements are noisy. Second,  road users' intentions remain ambiguous until they are in close proximity. Third, predicted trajectories branch across maneuver modes that demand fundamentally different responses from the AVs. Finally, the time window to plan safely is significantly compressed~\cite{schwarting2018planning}.

\begin{figure}[t]
    \centering
    \includegraphics[width=0.85\linewidth]{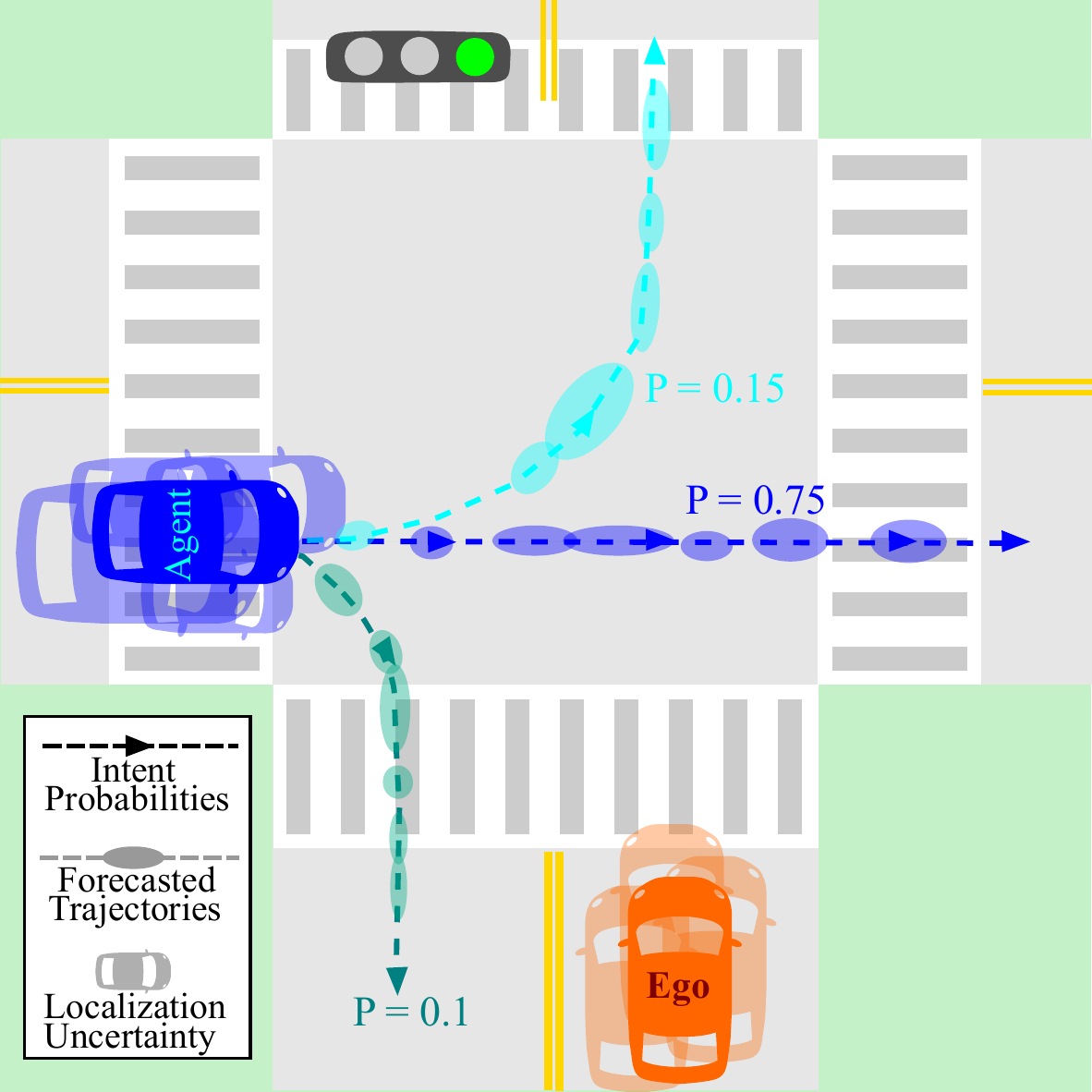}
    \caption{Multiple uncertainty sources in a two-vehicle intersection encounter. The agent's discrete intent is unknown, its forecasted trajectories accumulate covariance over the prediction horizon, and both vehicles carry localization uncertainty. Each source independently degrades collision avoidance if neglected (Table II), motivating a pipeline that preserves all of them as structured quantities across module boundaries.}
    \label{fig:overview}
    \vspace{-1.75em}
\end{figure}

To safely navigate an intersection, the planning algorithm must satisfy two requirements: (1) collision avoidance guarantees under uncertainty and (2) computational feasibility in real time. 
Deterministic MPC planners~\cite{nair2022collision} are computationally efficient but assume predicted trajectories as known, and lack the mechanism to ensure safety when an agent's true intent deviates from the assumed motion.
Chance-constrained formulations~\cite{zhu2019chance, schuurmans2023safe, bonzanini2022pac} address this by embedding probabilistic safety bounds in the optimization, but the nonlinearity of joint chance constraints over multi-step horizons frequently causes solver infeasibility to spike as uncertainty grows, degrading solution quality precisely when optimality is most critical. End-to-end learning methods~\cite{hu2023uniad,chen2024e2e_survey} bypass explicit uncertainty modeling altogether by absorbing it into opaque neural network parameters that offer neither interpretable safety margins nor formal feasibility guarantees~\cite{atakishiyev2024xai}.

These limitations point to a critical gap: existing frameworks lack a principled mechanism for propagating multi-modal uncertainty from its origin in perception to its point of use in planning and guaranteeing safety while maintaining real-time tractability. 
We address this with a modular perception-prediction-planning pipeline that propagates uncertainty in intent, trajectory forecasting, and localization (as shown in Fig.~\ref{fig:overview}) as structured, interpretable quantities across every module boundary. An Interacting Multiple Model (IMM) filter \cite{blom1988imm} jointly estimates the intent and kinematic state distributions of other road users (agents). Downstream modules use these estimates to forecast per-intent trajectories and compute collision probabilities that reflect the compounded uncertainty. A decision-making module then constructs convex safe sets from these probabilities, within which a differentially flat~\cite{fliess1995flatness} MPC planner operates with guaranteed solver feasibility at every cycle. A second order Control Barrier Function~\cite{ames2019cbf} safety filter enforces forward invariance of the safe set, providing a final guarantee that the executed trajectory respects the probabilistic safety bounds.

Our contributions are:
\begin{enumerate}
    \item \textbf{Structured uncertainty propagation across module boundaries.} We design a modular pipeline in which intent probabilities, forecast covariances, and localization error are propagated as explicit, interpretable quantities from perception through prediction to planning, enabling every downstream module to condition its output on the full upstream uncertainty quantification.
    \item \textbf{Concurrent safety and feasibility guarantees.} Our pipeline converts the uncertainties into collision probabilities upstream, and uses them to construct a probabilistically safe convex set to ensure collision avoidance.
    The convexity of this safe set ensures the feasibility of the planner and safety filter QPs at each cycle. 

    \item \textbf{Extensive validation and hardware deployment.} 
    We validate our pipeline through three complementary studies: (i)~a Monte Carlo benchmark against deterministic and chance-constrained MPC baselines, demonstrating simultaneous 0\% collision and 0\% infeasibility rates; (ii)~an ablation study that independently removes each uncertainty source, confirming that intent estimation, trajectory forecasting, and localization covariance are each necessary for collision avoidance while the feasibility guarantee holds structurally across all ablations; and (iii)~hardware deployment on Clearpath Jackal UGVs, verifying that both guarantees transfer to physical platforms under real sensing uncertainty.
\end{enumerate}

\section{Related Works}
\label{sec:related_works}

Planners for autonomous navigation in dynamic environments differ in how they represent and act on prediction uncertainty. We organize existing approaches into five classes and evaluate each against the two requirements identified in Section~\ref{sec:Introduction}: interpretable collision avoidance guarantees and real-time computational feasibility.

\textit{Robust and set-based methods}~\cite{althoff2016set} compute reachable sets that enclose all possible future behaviors and plan to avoid the entire set.
This guarantees collision avoidance by construction, but all maneuver intents collapse into a single over-approximation regardless of their likelihoods, which can lead to freezing behavior or numerical infeasibility.

\textit{Chance-constrained MPCs} (CC-MPCs)~\cite{zhu2019chance, schuurmans2023safe, bonzanini2022pac} replace worst-case avoidance with probabilistic safety bounds embedded directly as constraints in the trajectory optimization.
This reduces conservatism relative to set-based planning, but the required clearance from obstacles grows with prediction uncertainty, which compounds along the horizon; in narrow corridors or regions of high uncertainty, this clearance can exceed the available road space and cause solver infeasibility.

\textit{Sampling-based} approaches such as~\cite{degroot2025scenario} bound the collision probability over the full planning horizon by sampling obstacle trajectories from the predicted distribution and enforcing collision avoidance for each sample.
This yields tighter collision probability bounds than per-timestep CC-MPC, but more robust guarantees require more samples and therefore more computation, leading to a tradeoff between the strength of the safety guarantee and real-time feasibility.

\textit{POMDP planners}~\cite{hubmann2018pomdp, sunberg2022pomdp} maintain belief distributions over latent driver intent, coupling state estimation and trajectory planning in a single optimization over belief space.
Despite being the most principled treatment of intent ambiguity, the belief tree grows exponentially with the planning horizon. This leads to approximations of coarser action discretizations or shorter horizons that limit both the real-time applicability and the quality of the safety guarantee.

\textit{End-to-end learned pipelines}~\cite{hu2023uniad,
chen2024e2e_survey} train perception, prediction, and planning jointly, absorbing uncertainty into shared neural network parameters. The planner has no explicit uncertainty representation to reason over, so safety margins can neither be inspected nor formally guaranteed, and solver feasibility is replaced by a learned policy with no completeness assurance~\cite{atakishiyev2024xai}. Neither requirement is satisfied.

Across all five paradigms, a common limitation emerges: multi-modal uncertainties (\textit{e.g.} intent probabilities, forecasted trajectories, and localization errors) are either handled entirely within the planner, where it forces a tradeoff between safety and feasibility, or absorbed into opaque learned representations that provide neither guarantee. No existing framework propagates these quantities as structured, interpretable inputs across every module boundary while guaranteeing both collision avoidance and planner feasibility. This is the gap our work addresses.

\begin{figure*}[t]
    \centering
    \hspace{2.25em}  
\includegraphics[width=0.875\textwidth, height=4.25cm, keepaspectratio=false]{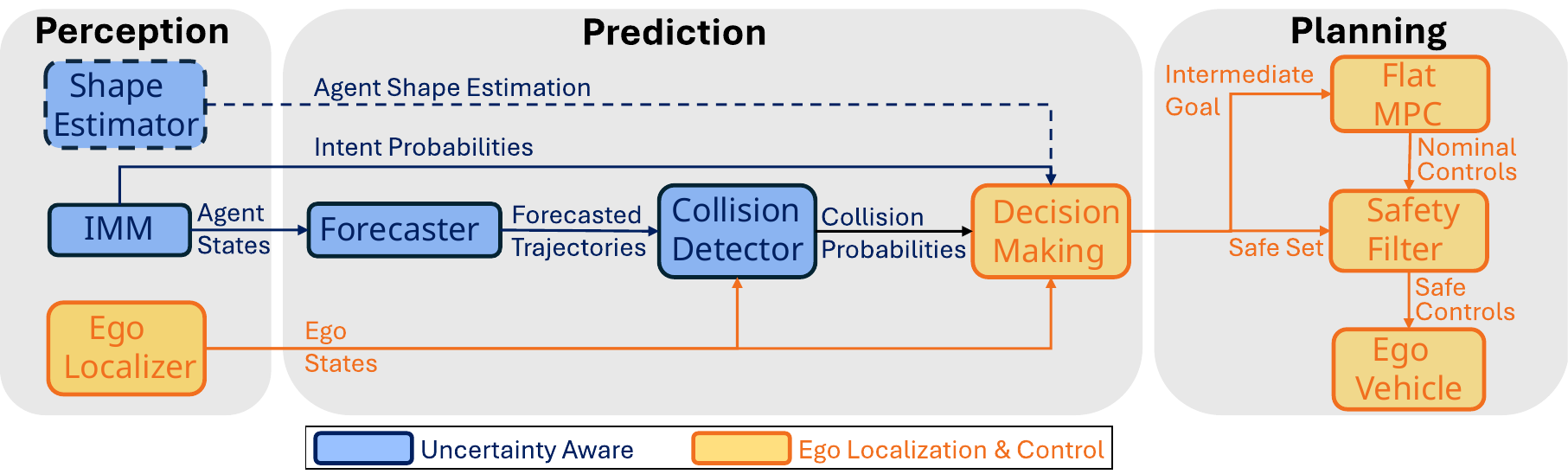}
    \caption{Pipeline Architecture Overview: The IMM estimates the agent's intent probabilities and states. The forecaster produces the forecasted trajectories. The collision detector combines these forecasted trajectories with the ego localizer's state estimates to compute per-intent collision probabilities. The decision maker converts these probabilities into a convex safe set, within which the flat MPC plans and the CBF safety filter enforces forward invariance.}
    \label{fig:pipeline}
    \vspace{-1.25em}
\end{figure*}

\section{Methods}

In this section, we define how each source of \textit{\textbf{Uncertainty}} is \textit{\textbf{Modeled}} and detail how the \textit{\textbf{Integrated Pipeline}} (Fig.~\ref{fig:pipeline}) propagates these uncertainties to produce real-time probabilistic collision avoidance and feasibility guarantees. 
\vspace{-0.5em}
\subsection{Uncertainty Modeling}
\subsubsection{Agent Intent \& Tracking}
An IMM estimator is used to track agent vehicles and estimate their intents. It is a multiple model state estimator that maintains two probability distributions over time: a discrete distribution over modes and a continuous distribution over the states of the tracked vehicle, where each mode represents one of the possible intents of the tracked vehicle. 

At any given timestep $k$, the estimator assumes the tracked vehicle obeys one of the modes defined by a unique model $M^j \in \mathcal{M} \triangleq \{M^1,...,M^r\}$, and that it can switch between modes. 
The mode set $\mathcal{M}$ is derived from road geometry, with one mode per feasible maneuver; for example, in a four-way intersection, $\mathcal{M} \triangleq \{\text{keep straight},\, \text{turn left},\, \text{turn right}\}$. 
The estimator runs $r$ filters (\textit{e.g.}, Extended Kalman Filters~\cite{gelb1974ekf}) in parallel, one for each model, and recursively updates the state estimate and the model probability. At timestep $k$, the probability that the current mode is $M^j$ is:
\vspace{-0.5em}
\begin{equation}
    \mu^j_k = \frac{1}{c} \Lambda^{j}_k \bar{c}^j \;\;\; j = 1,\ldots,r
\end{equation}
\noindent where $\Lambda^{j}_k$ is the likelihood of the measurement $\mathbf{z}_k$ under model $M^j$, $\bar{c}^j$ is the prior probability of $M^j$ before receiving measurement $\mathbf{z}_k$, and $c$ is a normalizing constant. Using mode probabilities $\mu^{j}_k$, the state estimate is updated as,
\vspace{-0.75em}
\begin{align}
&\hat{\mathbf{x}}_{k|k} = \sum_{j=1}^{r} \hat{\mathbf{x}}^j_{k|k} \mu^j_k&&\\
&P_{k|k} = \sum_{j=1}^{r} \mu^j_k \bigl\{ P^j_{k|k} + [\hat{\mathbf{x}}^j_{k|k} - \hat{\mathbf{x}}_{k|k}][\hat{\mathbf{x}}^j_{k|k} - \hat{\mathbf{x}}_{k|k}]^\top \bigr\}&&
\end{align}
\vspace{-0.25em}
\noindent where $\{\hat{\mathbf{x}}^j_{k|k}, P^j_{k|k}\}$ is the state estimate from the $j$-th filter with estimates conditioned on the current mode $M^j$.

By jointly estimating the
discrete intent and the continuous state distributions, the IMM provides downstream
modules with structured uncertainty representations: state covariance $P_{k|k}$ and intent probabilities $\boldsymbol{\mu}_k$.


\subsubsection{Agent Trajectory Forecasting}
\label{sec:forecaster}
For each individual intent mode $M^j \in \mathcal{M}$, the forecaster propagates the IMM state estimate $\hat{\mathbf{x}}_{k|k}$ and covariance $P_{k|k}$ forward over a horizon $T$. We use a Sigma Point Filter (SPF) for this, due to its ease in working with nonlinear kinematics. More specifically, the SPF uses a dynamics model with a lateral lane keeping controller and speed controller to ensure the agent follows the road; this closed-loop model has the added advantage of predicting covariance errors that are smaller, do not grow unbounded, and are aligned with the road.

\begin{figure*}[t]
    \centering
\includegraphics[width=0.9\textwidth]{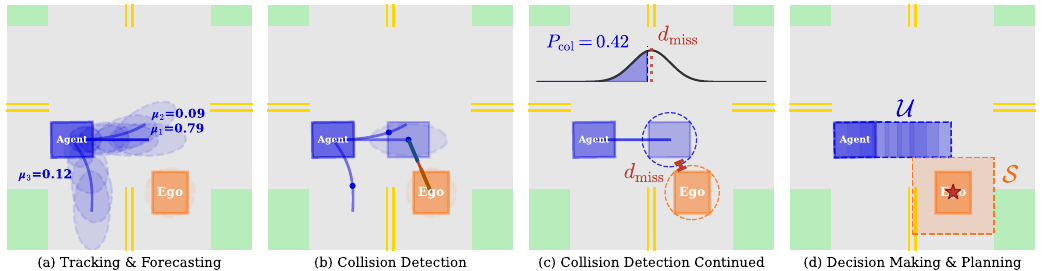}
    \vspace{-1em}
    \caption{Pipeline snapshot for a two-vehicle intersection encounter.
(a)~The IMM produces per-intent mode probabilities $\mu^j$; the forecaster propagates each into a trajectory (blue) with compounded covariance ellipses.
(b)~The collision detector finds the closest OBB approach $n^{*}$ (blue markers) along each forecast.
(c)~Combined covariance $\Sigma$ is projected onto the separation direction to yield $P_{\text{col}}^{j}$ analytically.
(d)~The decision maker sweeps the agent's OBB along the most likely forecast to form the unsafe region $\mathcal{U}$ and constructs the convex safe set $\mathcal{S}$ and determines the intermediate goal pose (shown as a red star) for the downstream planner and safety filter.}
    \label{fig:methods_fig}
    \vspace{-1.5em}
\end{figure*}

At each step $k$, $2n_x{+}1$ sigma points are generated
\begin{equation}
\chi^s_{k|k} \in \{\hat{\mathbf{x}}_{k|k},\; \hat{\mathbf{x}}_{k|k} \pm n_\sigma \mathbf{L}_k\}, \quad P_{k|k} = \mathbf{L}_k \mathbf{L}_k^\top
    \label{eq:sigma_pts}
\end{equation} 
and propagated through a kinematic bicycle model where the agent is assumed to track the maneuver path for intent $j$:
\begin{align}
    \theta^j_{k+1} &= \theta^j_k + \tfrac{v^j_k}{L}\sin(\delta^j_k)\,\Delta t, \label{eq:dyn_theta}\\
    \mathbf{p}^j_{k+1} &= \mathbf{p}^j_k + v^j_k
    \begin{bmatrix} \cos(\theta^j_{k+1}{+}\delta^j_k) \\ \sin(\theta^j_{k+1}{+}\delta^j_k) \end{bmatrix}\Delta t, \label{eq:dyn_pos}\\
    v^j_{k+1} &= v^j_k + a^j_k\,\Delta t, \label{eq:dyn_vel}
\end{align}
where $L$ is the wheelbase, $\delta_k^j$ is the steering input tracking the maneuver path, and $a_k^j$ is a proportional speed correction.
\vspace{-0.3em}
The predicted mean and covariance for each mode $j$ are:
\begin{equation}
    \hat{\mathbf{x}}^j_{k+1|k} = \sum_s W_m^s \, \boldsymbol{\chi}_{k+1|k}^s, \label{eq:ut_mean}
\end{equation}
\begin{equation}
    \hat{P}^j_{k+1|k} = \sum_s W_c^s \, \tilde{\boldsymbol{\chi}}^{s,j} \bigl(\tilde{\boldsymbol{\chi}}^{s,j}\bigr)^{\!\top} + Q,
    \label{eq:ut_cov}
\end{equation}
where $\tilde{\boldsymbol{\chi}}^{s,j} = \boldsymbol{\chi}_{k+1|k}^s - \hat{\mathbf{x}}^j_{k+1|k}$ and $Q$ is the additive process noise. 
The output is a per-intent trajectory $\{\hat{\mathbf{x}}^j_{k+n|k},\, \hat{P}^j_{k+n|k}\}_{n=1}^{N_T}$, where $N_T = T/\Delta t$ is the number of forecast steps over horizon $T$. This timestamped sequence of Gaussian state distributions allows collision probabilities to be evaluated analytically, as shown in Fig.~\ref{fig:methods_fig} (a-c).

\subsubsection{Ego Localization}
\label{sec:ego_loc}
The ego vehicle filters position measurements through an Extended Kalman Filter to estimate its pose $\hat{\mathbf{x}}^e_{k|k} = [x^e, y^e, \theta^e]^\top$ and position covariance $P^e_{k|k} \in \mathbb{R}^{2 \times 2}$. 
At each forecast timestep $n$, the collision detector combines both agent and ego localization uncertainties as $\Sigma =P^e_{k|k} + \hat{P}^j_{k+n|k}$. 
Therefore, noisier ego localization raises the estimated collision probabilities across all intent modes and leads to more cautious driving behavior.

\subsection{Integrated Pipeline}
The Collision Detection module converts the upstream uncertainties (agent intents, forecasted trajectories, localization errors) into scalar collision probabilities. The Decision Making module constructs a convex safe set for the planner and safety filter based on these probabilities. 
\subsubsection{Collision Detection}
\label{sec:collision_detection}
The collision detector converts the per-intent forecasted
trajectories $\{\hat{\mathbf{x}}^j_{k+n|k},\, \hat{P}^j_{k+n|k}\}_{n=1}^{N_T}$, the ego pose estimate $\hat{\mathbf{x}}^e_{k|k}$ with covariance $P^e_{k|k}$, and the vehicle geometry into a scalar collision probability for each intent.

An Oriented Bounding Box (OBB) is a rectangle oriented along the vehicle heading $\theta$ and parametrized by known half-extents $(e_0, e_1)$, half-length and half-width $(L/2,\, W/2)$. The full box spans $L \times M$.
The half-extents are calculated from known vehicle dimensions or estimated using vision, as shown in~\cite{monica2020shape}.

At the current timestep $k$, the ego and agent are represented as OBBs centered at their respective state estimate means ($\hat{\mathbf{x}}^e_{k|k}$ and $\hat{\mathbf{x}}_{k|k}$) with covariances ($P^e_{k|k}$ and $\hat{P}_{k|k}$). For each forecast timestep $n\in\{1,...,N_T\}$ for a given intent $j$, the forecasted agent pose is represented as an OBB centered at the mean $ \hat{\mathbf{x}}^j_{k+n|k} $ with covariance $\hat{P}^j_{k+n|k}$.


The relative position between the ego center at $k$ and the agent center at forecast timestep $n$ is a Gaussian random variable 
\begin{equation}
\begin{split}
    \Delta\mathbf{p}_{k,n} &\sim \mathcal{N}\!\left(\hat{\mathbf{x}}^j_{k+n|k} - \hat{\mathbf{x}}^e_{k|k},\; P^e_{k|k} + \hat{P}^j_{k+n|k}\right), \\
    & \quad n\in\{1,\dots,N_T\}
\end{split}
\label{eq:relative_position}
\end{equation}
The gap between the two OBBs along the center-to-center
direction $\mathbf{d}_{k,n} =(\hat{\mathbf{x}}^j_{k+n|k} - \hat{\mathbf{x}}^e_{k|k})/\|\hat{\mathbf{x}}^j_{k+n|k} - \hat{\mathbf{x}}^e_{k|k}\|$ is
\begin{equation}
\begin{split}
    d_{\mathrm{gap},k,n} &= \|\hat{\mathbf{x}}^j_{k+n|k} - \hat{\mathbf{x}}^e_{k|k}\| \\
    &\quad - r_{\mathrm{ego}}(\mathbf{d}_{k,n}) - r_{\mathrm{agent}}(\mathbf{d}_{k,n}),
\end{split}
\label{eq:obb_gap}
\end{equation}
where $r(\mathbf{d}) = e_0\,|\mathbf{a}_0 \cdot \mathbf{d}|
+ e_1\,|\mathbf{a}_1 \cdot \mathbf{d}|$ projects the vehicle OBB's body frame axes ($\mathbf{a}_0$ and $\mathbf{a}_1$) onto $\mathbf{d}$ and scales them by the half extents ($e_0$ and $e_1$). 
The timestep of closest approach is $n^* = \operatorname{argmin}_n\, d_{\mathrm{gap},k,n}$, shown in Fig.~\ref{fig:methods_fig}(b).
Because the projected OBB half-extents $r(\mathbf{d})$ vary with relative heading, the true point of closest approach can differ from the point of minimum center-to-center distance; incorporating vehicle geometry into this search allows the collision detector to more accurately calculate collision probabilities.

At $n^*$, the relative position $\Delta\mathbf{p}_{k,n^*}$ is Gaussian with combined
covariance $\Sigma = P^e_{k|k} + \hat{P}^j_{k+n^*|k}$. Projecting $\Sigma$ onto the center-to-center direction yields:
\begin{equation}
\begin{split}
    \sigma^2_{\mathrm{proj}} &= \mathbf{d}_{k,n^*}^\top \Sigma\, \mathbf{d}_{k,n^*}, \\
    d^j_{\mathrm{miss}} &= \|\hat{\mathbf{x}}^j_{k+n^*|k} - \hat{\mathbf{x}}^e_{k|k}\| - r_{\mathrm{col}},
\end{split}
\label{eq:proj}
\end{equation}
where $r_{\mathrm{col}} = r_{\mathrm{ego}} + r_{\mathrm{agent}}$
is the sum of bounding-circle radii that enclose each OBB at any orientation, so that the resulting collision
probability upperbounds the true risk.

The per-intent collision probability is then calculated as
\begin{equation}
    P^j_{\mathrm{col}} = 1 - \Phi\!\left(
        \frac{d^j_{\mathrm{miss}}}{\sigma_{\mathrm{proj}}}
    \right),
    \label{eq:pcol}
\end{equation}
where $\Phi$ is the standard normal CDF. The collision probabilities for all intents $\{P^j_{\mathrm{col}}\}_{M^j \in \mathcal{M}}$ are passed to the decision-making module to construct the safe set and determine the goal pose for the planner and safety filter.

\subsubsection{Decision Making}
\label{sec:decision_making}

The decision-making module converts upstream collision probabilities into a convex safe set and an intermediate goal pose that the downstream planner and safety filter can act on directly.
The decision-making module acts on the most likely intent $j^* = \arg\max_j \mu^j_k$ rather than a weighted combination $\sum_j \mu^j_k P^j_{\mathrm{col}}$.
A weighted sum can underestimate risk when a lower probability intent carries a high collision probability.
When intent probabilities are close (\textit{e.g.}, $\mathbf{\mu}_k = (0.45,0.5,0.05)$), the IMM's covariance $P_{k|k}$ grows, resulting in a larger $\Sigma$, higher $P^{j^*}_{\mathrm{col}}$, and a more conservative safe set.
Algorithm~\ref{alg:decision} summarizes the procedure. 
We define the safe set $\mathcal{S}$ as an Axis Aligned Bounding Box (AABB), a rectangular (convex) region that is aligned with the lane geometry. The lane geometry is defined by $\mathcal{L} = [x_L^-, x_L^+] \times [y_L^-, y_L^+]$, where $x_L^-$ and $x_L^+$ are the lateral lane bounds and $y_L^-$ and $y_L^+$ are the longitudinal lane entry and exit respectively.
\vspace{-0.75em}
\begin{algorithm}[h]
\small   
\caption{Decision Making}\label{alg:decision}
\begin{algorithmic}[1]
\State \textbf{Inputs:} 
\Statex Intent probabilities $\boldsymbol{\mu}_k$,
\Statex Per-intent collision probabilities $\{P^j_{\mathrm{col}}\}_{j=1}^{r}$,
\Statex Forecasted trajectories $\{\hat{\mathbf{x}}^{j}_{k+n|k}\}_{n=1}^{N_T}$,
\Statex Ego lane bounds $\mathcal{L}$,
\Statex Threshold $\epsilon$
\State\textbf{Outputs: }
\Statex Safe set $\mathcal{S}$
\Statex Intermediate goal pose $\zeta_{goal}$
\State $j^* \gets \arg\max_j \mu^j_k$
\If{$P^{j^*}_{\mathrm{col}} \leq \epsilon$}
    \State $\mathcal{S} \gets$ $\mathcal{L}$
 \Else
      \State $\mathcal{U} \gets \bigcup_{n} \mathrm{OBB}(\hat{\mathbf{x}}^{j^*}_{k+n|k})$
  $\triangleright$ sweep agent OBBs
      \If{$y_e < y_a$} 
      $\triangleright$ ego has not reached $\mathcal{U}$: yield
          \State $y_\text{safe} \gets \min_{(x,y)\,\in\,\mathcal{L} \cap \mathcal{U}} y$
          \State $\mathcal{S} \gets [x_{\mathcal{L}}^{-},\, x_{\mathcal{L}}^{+}] \times [y_e -
  L,\; y_\text{safe}]$
      \Else \quad $\triangleright$ $\mathcal{U}$ behind ego: proceed to exit $\mathcal{U}$
          \State $\mathcal{S} \gets \mathcal{L}$
      \EndIf
  \EndIf
\State $\zeta_{\mathrm{goal}} \gets$ intermediate goal pose within $\mathcal{S}$
\end{algorithmic}
\end{algorithm}
\vspace{-0.75em}
Fig.~\ref{fig:methods_fig} illustrates a full pipeline cycle of a high-risk scenario for a two-vehicle encounter. The most likely intent is keep-straight ($\mu^1 = 0.79$), with collision probability $P^{j^*}_{\mathrm{col}} = 0.42 > \epsilon = 0.01$. The agent's OBB footprint is swept along the keep-straight forecast to form the unsafe region $\mathcal{U}$, and the safe set $\mathcal{S}$ is constructed as the largest AABB within the ego lane that excludes $\mathcal{U}$:
\vspace{-0.5em}
\begin{equation}
    \mathcal{S} = [x_L^-, x_L^+] \times [y_{\min}, y_{\max}] = \bigcap_{m=1}^{4} H_m,
    \vspace{-0.75em}
\end{equation}
where $y_{\min}, y_{\max}$ are set by Algorithm~\ref{alg:decision}: 
either $[y_L^-, y_L^+]$ when $P^{j^*}_{\mathrm{col}} \leq \epsilon$, or $[y_e - L,\ y_{\mathrm{safe}}]$ in the yield case. Each $H_m$ is a half-plane
 
By definition of convexity, the straight line path connecting two points (the ego's current pose and the intermediate goal pose) lies entirely within the convex set. Therefore, the safe set's convexity ensures the recursive feasibility of the MPC planner.
Since the safe set $\mathcal{S}$ is constructed to be probabilistically safe, any trajectory contained within it inherits the safety bounds.
Furthermore, the AABB structure of $\mathcal{S}$ yields a set of linear inequality constraints for the CBF safety filter, preserving the QP's tractability and ensuring forward invariance.  

\subsubsection{Planner and Safety Filter}
\label{sec:planner}
The planner generates a trajectory toward the intermediate goal pose $\zeta_{goal}$ and the safety filter ensures the ego stays within the safe set $\mathcal{S}$. 

Differential flatness is an algebraic property that transforms the system dynamics to be linear, controllable, and time-invariant~\cite{fliess1995flatness}. 
It maps the nonlinear vehicle dynamics to a double-integrator in flat-output space $\mathbf{\zeta} = [x,\,y,\,v_x,\,v_y]^\top$ with control input $\mathbf{u} = [a_x,\,a_y]^\top$. 
The planner objective function is
\begin{equation}
    \min_{\zeta,\mathbf{u}}\;
    \sum_{k=0}^{N-1}\!\bigl[
        \|\zeta_k - \zeta_{\mathrm{goal}}\|^2_Q + \|\mathbf{u}_k\|^2_R
    \bigr]
    + \|\zeta_N - \zeta_{\mathrm{goal}}\|^2_P
    \label{eq:mpc}
\end{equation}
subject to linear dynamics $\zeta_{k+1} = A\zeta_k + B\mathbf{u}_k$, speed and acceleration limits, and the initial condition $\zeta_0 = \zeta_{\mathrm{current}}$. 
The linear dynamics and quadratic cost yield a small optimization solved in ${\sim}5$\,ms.

Control Barrier Functions are a natural choice for enforcing the safe set because Nagumo's theorem guarantees forward invariance: if the CBF condition is satisfied at every timestep, the state remains inside $\mathcal{S}$ for all future time~\cite{Blanchini1999SetII}.
For each boundary $b \in \{x_{\min}, x_{\max}, y_{\min}, y_{\max}\}$, a barrier function $h_b(\zeta,
\mathbf{u})$ is defined such that $h_b \geq 0$ keeps the state inside $\mathcal{S}$. 
For example, for the upper $y$-boundary, the barrier function $h_{y_{\max}}$ encodes the constraint $y \leq y_{\max}$, and its
second-order CBF condition is
\begin{equation}
\begin{split}
    h_{y_{\max}}(\zeta, \mathbf{u})
    &= -a_y - (\lambda_1 + \lambda_2)\,v_y \\
    &\quad + \lambda_1 \lambda_2\,(y_{\max} - y)
    \;\geq\; 0,
\end{split}
\label{eq:cbf_h}
\end{equation}
where $\lambda_1, \lambda_2 > 0$ are tunable parameters; the remaining three boundaries follow the same formulation.

For the double integrator, each CBF condition is \emph{linear in the acceleration input},
yielding the following QP:
\begin{equation}
\begin{split}
    &\min_{\mathbf{u}_{\mathrm{safe}}}\;
    \|\mathbf{u}_{\mathrm{safe}} - \mathbf{u}_{\mathrm{ref}}\|^2 \\
    &\text{s.t.}\quad
    h_b(\zeta_{\mathrm{current}},\, \mathbf{u}_{\mathrm{safe}}) \geq 0
    \;\;\forall\, b,
\end{split}
\label{eq:cbf_qp}
\end{equation}
which minimally modifies the planner's reference acceleration
to satisfy all four boundary constraints. The safe flat-space controls $\mathbf{u}_{\mathrm{safe}} = [a_x,\, a_y]^\top$ are then mapped back to unicycle controls via the flatness relations $v = \sqrt{v_x^2 + v_y^2}$ and $\omega = (v_x a_y - v_y a_x)/v^2$, yielding the linear and angular velocities applied to the system.

Together, the pipeline propagates structured uncertainties from perception to planning, providing interpretable collision avoidance guarantees while maintaining computational tractability.

\section{Validation and Results}
\vspace{-0.25em}
\subsection{Monte Carlo Simulation Results}
\vspace{-0.25em}
\begin{table*}[t]
\centering
\small
\setlength{\tabcolsep}{4pt}
\caption{Monte Carlo Analysis Results}
\label{tab:mc_comparison}
\vspace{-0.5em}
\begin{tabular}{l l r r r c c c r c c}
\toprule
Planner & Scenario & $N$ & Coll.\% & Succ.\% & OBB Dist (m) & Goal Time (s) & Norm.\ Time & Infeas.\% & QP (ms) & Cycle (ms) \\
\midrule
Ours & Keep Straight & 35 & 0.0 & 100.0 & $0.18 \pm 0.03$ & $13.07 \pm 0.73$ & $2.09 \pm 0.12$ & 0.0 & $4.7 \pm 2.4$ & $12.4 \pm 2.5$ \\
Ours & Left Turn & 37 & 0.0 & 100.0 & $0.26 \pm 0.03$ & $12.96 \pm 0.94$ & $2.07 \pm 0.15$ & 0.0 & $5.4 \pm 3.4$ & $14.2 \pm 3.4$ \\
Ours & Right Turn & 28 & 0.0 & 100.0 & $0.30 \pm 0.01$ & $9.16 \pm 0.55$ & $1.47 \pm 0.09$ & 0.0 & $5.2 \pm 3.0$ & $11.0 \pm 3.0$ \\
\textbf{Ours} & \textbf{Overall} & 100 & 0.0 & 100.0 & $0.25 \pm 0.06$ & $11.94 \pm 1.90$ & $1.91 \pm 0.30$ & 0.0 & $5.1 \pm 2.9$ & $12.7 \pm 3.3$ \\

\midrule
MPC & Keep Straight & 35 & 2.9 & 97.1 & $0.08 \pm 0.03$ & $7.70 \pm 0.71$ & $1.23 \pm 0.11$ & 0.0 & $18.4 \pm 3.2$ & $18.4 \pm 3.2$ \\
MPC & Left Turn & 37 & 94.6 & 5.4 & $0.10 \pm 0.00$ & $11.20 \pm 0.28$ & $1.79 \pm 0.05$ & 0.0 & $23.5 \pm 3.4$ & $23.5 \pm 3.4$ \\
MPC & Right Turn & 28 & 0.0 & 100.0 & $0.50 \pm 0.03$ & $7.40 \pm 0.20$ & $1.18 \pm 0.03$ & 0.0 & $17.2 \pm 2.8$ & $17.2 \pm 2.8$ \\
\textbf{MPC} & \textbf{Overall} & 100 & 36.0 & 64.0 & $0.26 \pm 0.21$ & $7.68 \pm 0.84$ & $1.23 \pm 0.13$ & 0.0 & $18.0 \pm 3.2$ & $18.0 \pm 3.2$ \\
\midrule
CC-MPC & Keep Straight & 35 & 11.4 & 68.6 & $0.23 \pm 0.10$ & $13.10 \pm 1.76$ & $2.10 \pm 0.28$ & 19.2 & $19.9 \pm 7.0$ & $24.9 \pm 7.1$ \\
CC-MPC & Left Turn & 37 & 16.2 & 16.2 & $0.41 \pm 0.02$ & $14.05 \pm 0.56$ & $2.25 \pm 0.09$ & 17.1 & $17.6 \pm 1.5$ & $23.4 \pm 1.5$ \\
CC-MPC & Right Turn & 28 & 0.0 & 100.0 & $0.49 \pm 0.08$ & $10.47 \pm 0.96$ & $1.67 \pm 0.15$ & 4.0 & $16.6 \pm 3.1$ & $20.4 \pm 3.1$ \\
\textbf{CC-MPC} & \textbf{Overall} & 100 & 10.0 & 58.0 & $0.37 \pm 0.15$ & $11.93 \pm 1.95$ & $1.91 \pm 0.31$ & 14.2 & $18.1 \pm 5.2$ & $22.6 \pm 5.5$ \\
\midrule
Ours & Three-Way & 100 & 0.0 & 99.0 & $0.29 \pm 0.02$ & $11.25 \pm 2.27$ & $1.80 \pm 0.36$ & 0.0 & $3.0 \pm 0.1$ & $7.2 \pm 0.9$ \\
Ours & Multi-Agent & 100 & 0.0 & 100.0 & $0.29 \pm 0.07$ & $17.98 \pm 0.80$ & $2.88 \pm 0.13$ & 0.0 & $3.0 \pm 0.2$ & $15.3 \pm 1.1$ \\
\bottomrule
\end{tabular}
\vspace{-1.75em}
\end{table*}
We simulate 100 Monte Carlo trials in a four-way intersection with two vehicles. 
The agent's intent is sampled randomly from $\mathcal{M}$. Gaussian noise is injected into the pose measurements for both the agent and the ego.
The collision detector uses a probability threshold of $\epsilon = 0.01$ to flag high-risk events.

We compare our pipeline against two baselines:

\begin{enumerate}
    \item \noindent\hangindent=1.5em\hangafter=1
    \textit{Deterministic MPC} uses a constant-velocity EKF to track the agent without intent estimation or trajectory forecasting. 
    The planner solves a differentially flat QP with a second-order CBF collision avoidance constraint.
    \item \noindent\hangindent=1.5em\hangafter=1
    \textit{CC-MPC} uses our IMM intent estimator and SPF trajectory forecaster, but embeds safety as per-timestep chance constraints requiring clearance $D + \Phi^{-1}(0.99)\,\sigma_{\mathrm{proj}}$ along the separation direction~\cite{zhu2019chance}.
\end{enumerate}

\noindent 
Both baselines fall back to their last feasible control when the solver becomes infeasible.
All three planning frameworks are implemented using CasADi IPOPT~\cite{andersson2019casadi}.

Additionally, we test our pipeline with two extensions: a three-way intersection as a different road topology and a multi-agent scenario of three agents in a sequential cue, whose intents are independent of each other.

We evaluate each planner on the following metrics. 

\begin{enumerate}
\item \noindent\hangindent=1.5em\hangafter=1
\textit{Coll.\%} is the percentage of trials that result in collisions (defined as an overlap between the ego and agent OBBs).

\item \noindent\hangindent=1.5em\hangafter=1
\textit{Succ.\%} is the percentage of trials in which the ego reaches its lane exit without a collision.

\item \noindent\hangindent=1.5em\hangafter=1
\textit{OBB Dist} is the minimum body-to-body clearance between the ego and agent OBBs throughout the trial.

\item \noindent\hangindent=1.5em\hangafter=1
\textit{Goal Time} is the time to reach the lane exit.

\item \noindent\hangindent=1.5em\hangafter=1
\textit{Norm.\ Time} is the time to reach the lane exit divided by an obstacle-free travel time.

\item \noindent\hangindent=1.5em\hangafter=1
\textit{Infeas.\%} is the percentage of planning cycles in which the solver returns no solution and the planner had to fall back to the last set of feasible control inputs. 

\item \noindent\hangindent=1.5em\hangafter=1
\textit{QP (ms)} reports the solver time for the planner and safety filter alone.

\item \noindent\hangindent=1.5em\hangafter=1
\textit{Cycle (ms)} reports the wall-clock time for the full pipeline cycle including perception, prediction, and decision making. Reporting both separates the cost of the upstream uncertainty modules from the path planner itself, showing that the computational overhead of structured uncertainty propagation is non-negligible but modest, and necessary.
\end{enumerate}
\noindent The distance and time metrics are only calculated for the trials in which the ego successfully reaches the lane exit without a collision.

Table~\ref{tab:mc_comparison} evaluates our approach against the two baselines and in two scenario extensions using these metrics.
Our pipeline achieves 0\% collisions across all 100 trials, with an 
OBB clearance of $0.25 \pm 0.06$\,m. The deterministic MPC collides in 36\% of trials (94.6\% in left-turn); this is primarily due to lacking intent estimation, which causes all predictions to be constant velocity and straight. The ego vehicle cannot brake in time for cases when the agent turns left into its lane.
The CC-MPC reduces collisions to 10\% but still fails in 11.4\% of keep-straight trials and 16.2\% of left-turn trials. These are primarily due to the ego and agent covariances compounding along the horizon, so the required clearance exceeds the lane width and makes the solver infeasible. 
In this case, the planner falls back to the last feasible control input, which causes collision.

Our pipeline achieves 0\% solver infeasibility across all 100 trials: the safe set always contains a collision-free path from the ego's current position.
The CC-MPC experiences 14.2\% infeasibility overall (17.1\% in left-turn), and only 58\% of its collision-free trials reach the goal, while the rest freeze from repeated infeasibilities.

Because collision probabilities and safe sets are resolved analytically upstream, the planner and safety filter in our pipeline both reduce to a small convex QP, completing the full pipeline cycle in $12.7 \pm 3.3$\,ms, 29\% faster than the deterministic MPC's $18.0 \pm 3.2$\,ms and 44\% faster than the CC-MPC's $22.6 \pm 5.5$\,ms.

Our pipeline generalizes to alternative scenarios: in the three-way intersection, it maintains 0\% collisions, with a cycle time of $7.2 \pm 0.9$ ms. 
One trial reports unreached goal, with a 6 cm offset from the goal against the 5cm threshold.
In the multi-agent extension, the pipeline maintains 0\% collisions and 0\% infeasibility, with a longer cycle time ($15.3 \pm 1.1$ ms) as it estimates each agent's intent individually.
\vspace{-0.5em}
\begin{table*}[b]
\centering
\small
\setlength{\tabcolsep}{3pt}
\vspace{-1em}
\caption{Ablation Study Results}
\label{tab:ablation}
\vspace{-0.5em}
\begin{tabular}{l l r r r c c c r c c}
\toprule
Planner & Scenario & $N$ & Coll.\% & Succ.\% & OBB Dist (m) & Goal Time (s) & Norm.\ Time & Infeas.\% & QP (ms) & Cycle (ms) \\
\midrule
Full & Keep Straight & 35 & 0.0 & 100.0 & $0.18 \pm 0.03$ & $13.07 \pm 0.73$ & $2.09 \pm 0.12$ & 0.0 & $4.7 \pm 2.4$ & $12.4 \pm 2.5$ \\
Full & Left Turn & 37 & 0.0 & 100.0 & $0.26 \pm 0.03$ & $12.96 \pm 0.94$ & $2.07 \pm 0.15$ & 0.0 & $5.4 \pm 3.4$ & $14.2 \pm 3.4$ \\
Full & Right Turn & 28 & 0.0 & 100.0 & $0.30 \pm 0.01$ & $9.16 \pm 0.55$ & $1.47 \pm 0.09$ & 0.0 & $5.2 \pm 3.0$ & $11.0 \pm 3.0$ \\
\textbf{Full} & \textbf{Overall} & 100 & 0.0 & 100.0 & $0.25 \pm 0.06$ & $11.94 \pm 1.90$ & $1.91 \pm 0.30$ & 0.0 & $5.1 \pm 2.9$ & $12.7 \pm 3.3$ \\
\midrule
No Intent & Keep Straight & 35 & 45.7 & 54.3 & $0.19 \pm 0.04$ & $12.87 \pm 0.51$ & $2.06 \pm 0.08$ & 0.0 & $3.0 \pm 0.1$ & $8.0 \pm 0.3$ \\
No Intent & Left Turn & 37 & 35.1 & 64.9 & $0.23 \pm 0.08$ & $12.71 \pm 1.36$ & $2.03 \pm 0.22$ & 0.0 & $2.9 \pm 0.1$ & $8.7 \pm 0.4$ \\
No Intent & Right Turn & 28 & 0.0 & 100.0 & $0.31 \pm 0.02$ & $8.90 \pm 1.23$ & $1.42 \pm 0.20$ & 0.0 & $3.1 \pm 0.1$ & $6.9 \pm 0.3$ \\
\textbf{No Intent} & \textbf{Overall} & 100 & 29.0 & 71.0 & $0.25 \pm 0.07$ & $11.25 \pm 2.22$ & $1.80 \pm 0.35$ & 0.0 & $3.0 \pm 0.1$ & $7.8 \pm 0.9$ \\
\midrule
No Forecasting & Keep Straight & 35 & 20.0 & 68.6 & $0.19 \pm 0.08$ & $13.38 \pm 0.54$ & $2.14 \pm 0.09$ & 0.0 & $4.4 \pm 1.6$ & $6.3 \pm 1.6$ \\
No Forecasting & Left Turn & 37 & 5.4 & 75.7 & $0.22 \pm 0.09$ & $13.02 \pm 0.90$ & $2.08 \pm 0.14$ & 0.0 & $4.8 \pm 3.3$ & $6.7 \pm 3.2$ \\
No Forecasting & Right Turn & 28 & 0.0 & 100.0 & $0.30 \pm 0.01$ & $9.36 \pm 0.45$ & $1.50 \pm 0.07$ & 0.0 & $5.3 \pm 4.0$ & $7.0 \pm 4.0$ \\
\textbf{No Forecasting} & \textbf{Overall} & 100 & 9.0 & 80.0 & $0.24 \pm 0.08$ & $11.85 \pm 1.96$ & $1.90 \pm 0.31$ & 0.0 & $4.9 \pm 3.2$ & $6.7 \pm 3.1$ \\
\midrule
No Localization & Keep Straight & 35 & 14.3 & 85.7 & $0.12 \pm 0.03$ & $12.59 \pm 0.76$ & $2.01 \pm 0.12$ & 0.0 & $2.8 \pm 0.1$ & $6.9 \pm 0.3$ \\
No Localization & Left Turn & 37 & 0.0 & 100.0 & $0.21 \pm 0.04$ & $13.20 \pm 0.80$ & $2.11 \pm 0.13$ & 0.0 & $2.8 \pm 0.1$ & $7.3 \pm 0.4$ \\
No Localization & Right Turn & 28 & 0.0 & 100.0 & $0.30 \pm 0.01$ & $8.77 \pm 0.57$ & $1.40 \pm 0.09$ & 0.0 & $2.9 \pm 0.1$ & $6.2 \pm 0.3$ \\
\textbf{No Localization} & \textbf{Overall} & 100 & 5.0 & 95.0 & $0.21 \pm 0.08$ & $11.70 \pm 2.05$ & $1.87 \pm 0.33$ & 0.0 & $2.8 \pm 0.1$ & $6.8 \pm 0.6$ \\
\bottomrule
\end{tabular}
\end{table*}
\begin{figure*}[b]
\vspace{-1em}
    \centering
    \hspace{.5em}    
    \includegraphics[width=0.8 \textwidth]{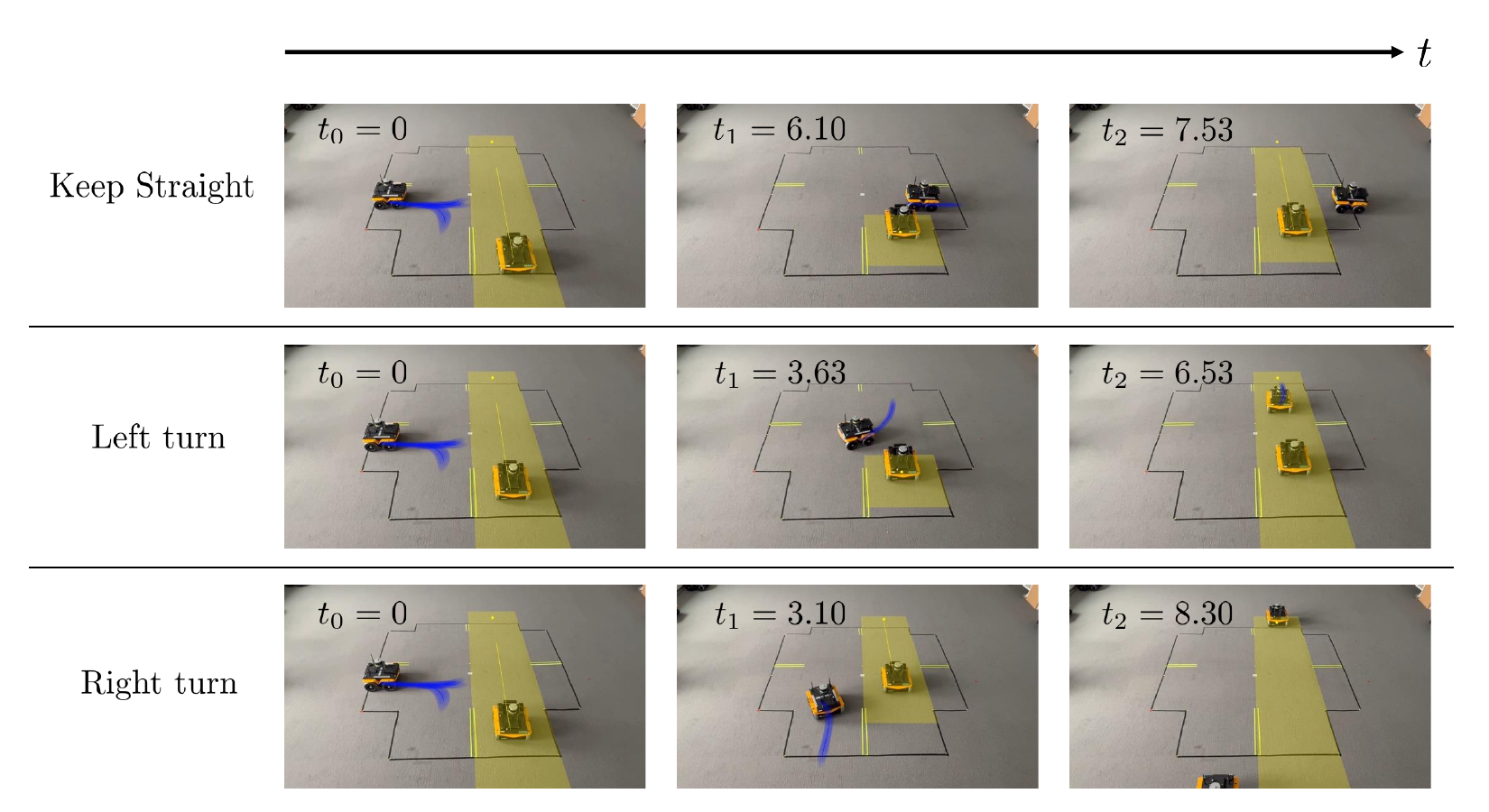}
    \vspace{-1.5em}
    \caption{
    Snapshots of three experiment scenarios at three different times: $t_0$ is the start time, $t_1$ is the time of highest collision probability, and $t_2$ is the time when the unsafe set no longer intersects with the lane. 
    The forecasted trajectories are shown in blue (lines and covariance ellipsoids).
    The ego's planned trajectory, goal point, and safe set are shown in yellow (as line, dot, and rectangle, respectively)
    }
    \label{fig:experiment}
\end{figure*}

\vspace{-1em}
\subsection{Ablation Study}
To isolate the contribution of each uncertainty source, we define three ablation conditions that each substitutes one uncertainty while leaving the rest of the pipeline intact.

\begin{enumerate}
    \item \noindent\hangindent=1.5em\hangafter=1
    \textit{No Intent} bypasses the IMM intent estimation by randomly assigning a fixed intent probability vector (one of $[1,0,0]$, $[0,1,0]$, or $[0,0,1]$), independent of the agent's true maneuver.
    The forecaster, collision detector, and decision maker operates normally but condition on an intent incorrect $66.7\%$ of the time.
    
    \item\noindent\hangindent=1.5em\hangafter=1
    \textit{No Forecasting} replaces the SPF-based per-intent forecaster with an EKF following a constant-velocity model, whose covariance grows linearly with time.
    The constant-velocity EKF extrapolates the agent's trajectory linearly from the current state, producing forecast errors that grow with the horizon, and significantly larger covariances whenever the agent turns.
    
    \item\noindent\hangindent=1.5em\hangafter=1
    \textit{No Localization} supplies near-perfect location measurements, causing the pipeline to underestimate the combined covariance $\Sigma$ used in collision detection. 
\end{enumerate}
Table~\ref{tab:ablation} summarizes the results of the ablation study. 
Removing intent estimation (No Intent) produces the steepest degradation: collisions rise to 29\%.
Collisions are concentrated in the keep straight and left turn scenarios, when the collision detection and decision making modules fail to construct a yielding safe set under the incorrect intent. 
Replacing the forecaster with a constant linear velocity model (No Forecasting) raises collisions to 9\%.
The straight-line forecast cannot anticipate turning maneuvers, and its linearly growing covariance misjudges collision timing even in the keep-straight scenario, where the predicted direction is approximately correct. 
Eliminating the agent and ego localization uncertainties by supplying near-perfect localization (No Localization) increases collisions to 5\%: without capturing the localization covariance, the pipeline underestimates collision risk, drives more aggressively, and leaves insufficient clearance for braking. 
These results confirm that agent intent, forecasting, and localization covariances are each independently necessary for collision avoidance.

All three ablation conditions maintain 0\% solver infeasibility, although the trajectory quality might degrade depending on the upstream uncertainty propagation. The full pipeline's additional ${\sim}6$~ms over the computationally cheaper ablation is a modest price for the localization, IMM, and uncertainty-aware forecaster that provide better collision avoidance performance.

\begin{figure*}[t]
    \centering
    \hspace{.5em}    
    \includegraphics[width=0.8\textwidth]{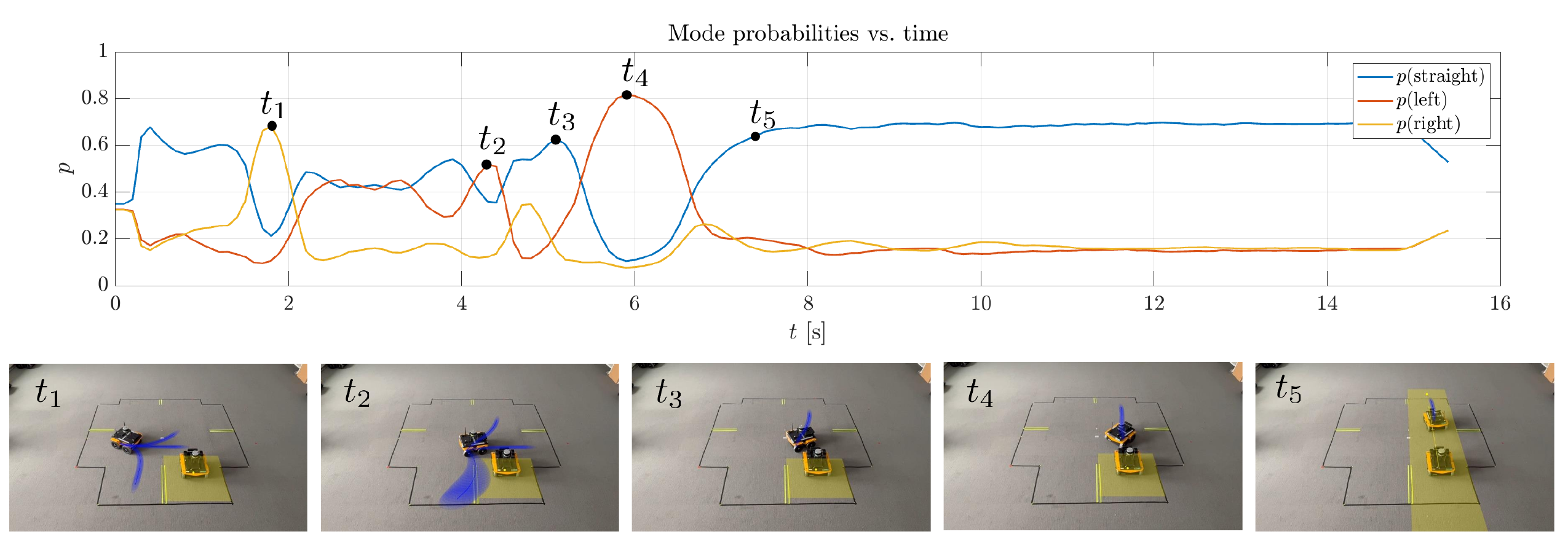}
    \vspace{-1.5em}
    \caption{The IMM estimator outputs oscillating intent probabilities as the agent follows a zig-zag trajectory not in $\mathcal{M}$. Five timesteps are shown: $t_1$---agent turns right; $t_2$---switches to left; $t_3$---switches to straight; $t_4$---commits to left; $t_5$---ego deems it safe to proceed.}
    \label{fig:ood}
    \vspace{-1.5em}
\end{figure*}
\vspace{-0.5em}
\subsection{Hardware Deployment} 
We deploy our pipeline on Clearpath Jackal UGVs in a 3m × 3m intersection testbed to validate its real-time viability on physical hardware. 
In all three scenarios, the ego robot is tasked with crossing the intersection safely in the presence of an agent robot whose pose and intent are uncertain.


A motion capture system provides $(x,y)$ measurements of both robots at 60 Hz with no additional noise added. The ego filters its own position measurements with an EKF (50 Hz) to estimate its pose and velocity. It filters the agent’s position measurements with an IMM estimator (10 Hz) that acts as both a tracker and an intent estimator, jointly estimating the agent’s pose, velocity, and intent. Downstream, the forecaster (25 Hz, 2.5s horizon), collision detector (20 Hz), and decision maker (50 Hz) propagate the uncertainties and construct the safe set and intermediate goal that feed into the planner (2 Hz) and safety filter (5 Hz). The collision detector uses a probability threshold of $\epsilon = 0.01$.

Across three experiment scenarios, we vary the agent's behavior (unknown to the ego) as the ego attempts to cross the intersection, as shown in Fig.~\ref{fig:experiment}.

In all three scenarios, the ego robot successfully navigates the intersection without collisions, mirroring the simulation results. In both the keep straight and left turn scenarios, our pipeline detected high collision probabilities and proactively contracted the safe sets to avoid collisions, maintaining a body-to-body clearance of  0.0869m and 0.2641m, and completion time of 13.16s and 12.9s respectively. In the right turn scenario, the agent's path did not cross the ego's lane, so the pipeline recognizes the low collision probability, and restores the full lane as the safe set early, with a body-to-body clearance of 0.6060m and time to completion of 7.46s. The results validate the collision avoidance guarantee under real hardware's time-varying sensing uncertainty. The feasibility guarantee also holds at every cycle across all three scenarios.

To further probe our pipeline's robustness against unmodeled behavior, we add a fourth scenario in which the agent follows a zig-zagging trajectory that does not correspond to any intent mode in $\mathcal{M}$. Because the IMM cannot explain the trajectory with a single mode, intent probabilities oscillate, keeping collision risk elevated and the safe set conservatively contracted throughout the maneuver, as shown in Fig.~\ref{fig:ood}. The ego successfully avoids collision with a minimum clearance of 0.2641m and completion time of 15.33s. This out-of-distribution experiment suggests that the pipeline degrades gracefully under model mismatch: the upstream uncertainty modules produce conservative estimates when no mode fits well, preserving the collision avoidance and feasibility properties observed in the nominal scenarios.

\section{Conclusion}
We presented an uncertainty-aware perception-prediction-planning pipeline that addresses two requirements for safe autonomous navigation: collision avoidance under uncertainty and real-time feasibility. 
By propagating intent probabilities, forecast covariances, localization errors, and vehicle geometry as structured quantities across module boundaries, our pipeline produces convex safe sets that adapt to the current level of uncertainty.
In 100 Monte Carlo trials, our pipeline achieves 0\% collisions and 0\% solver infeasibility at 12.7~ms per cycle, compared to 36\% collisions for the deterministic MPC and 10\% collisions with 14.2\% infeasibility for the chance-constrained MPC. The ablation study confirms that intent estimation, forecasting, and localization covariances are each independently necessary for the collision avoidance guarantee, while the feasibility guarantee holds structurally across all ablations. Hardware deployment on Clearpath Jackal UGVs validates that these guarantees transfer from simulation to physical platforms. In summary, our pipeline addresses the gap of simultaneously guaranteeing collision avoidance and feasibility in real time, and provides an interpretable, modular approach toward certifiably safe autonomous driving.

\bibliographystyle{IEEEtran}
\bibliography{08_references}

\end{document}